\documentclass[sigconf, screen, natbib=false]{acmart}

\AtBeginDocument{%
  }

\setcopyright{rightsretained}
\copyrightyear{2023}
\acmYear{2023}

\RequirePackage[
  datamodel=acmdatamodel,
  style=acmnumeric,
  ]{biblatex}

\begin{document}

\title{An Analysis of Dialogue Repair in Voice Assistants}

\author{Matthew Galbraith}
\email{matthew.galbraith@upf.edu}
\orcid{0000-0002-6299-2610}
\affiliation{%
  \institution{Universitat Pompeu Fabra}
  \city{Barcelona}
  \country{Spain}
  \postcode{08018}
}


\begin{abstract}
Spoken dialogue systems have transformed human-machine interaction by providing real-time responses to queries. However, misunderstandings between the user and system persist. This study explores the significance of interactional language in dialogue repair between virtual assistants and users by analyzing interactions with Google Assistant and Siri, focusing on their utilization and response to the other-initiated repair strategy "huh?" prevalent in human-human interaction. Findings reveal several assistant-generated strategies but an inability to replicate human-like repair strategies such as "huh?". English and Spanish user acceptability surveys show differences in users’ repair strategy preferences and assistant usage, with both similarities and disparities among the two surveyed languages. These results shed light on inequalities between interactional language in human-human interaction and human-machine interaction, underscoring the need for further research on the impact of interactional language in human-machine interaction in English and beyond. 
\end{abstract}

\ccsdesc[300]{Human-centered computing~User studies}
\ccsdesc[500]{Human-centered computing~Human computer interaction (HCI)}
\ccsdesc[500]{Human-centered computing~Usability testing}

\keywords{spoken dialogue systems, interactional language, dialogue repair, virtual assistants, human-machine interaction}

\maketitle

\section{Introduction}
\subsection{Spoken Dialogue Systems}
In recent years, spoken dialogue systems (SDS) ranging from popular voice assistants like Apple’s Siri and Google’s Google Assistant (two of the most popular SDS \cite{hoy_alexa_2018}) to emerging systems in smart homes, automobiles, and robotic personal companions have become an integral part of daily life, revolutionizing the way that we interact with technology and shaping the fundamental nature of human-machine interaction (HMI). The increasing ubiquity of voice-based technology has sparked a growing interest in understanding the inner workings of these systems and how to improve them, prompting the need for more comprehensive research. Moreover, it is crucial to expand this examination to encompass SDS functioning in languages other than English such as Spanish, which boasts a substantial number of speakers worldwide but can lack the necessary attention and resources to build robust linguistic data for complex systems \cite{gutierrez-fandino_maria_2022}.

\subsection{Interactional Language in HHI}
A hallmark of human-human interaction (HHI) is the use of interactional language which encompasses the complex dynamics and strategies employed by humans in conversation. Interactional language includes the linguistic, paralinguistic, and pragmatic features involved in conversations including turn-taking, politeness, repair mechanisms, and prosodic cues \cite{ginzburg_interactive_2012} \cite{levinson_interactional_2019} \cite{wiltschko_grammar_2021}. Within the realm of interactional language, an integral aspect known as other-initiated repair (OIR), which falls under the broader category of dialogue repair, occurs at a staggering rate of 1.4 instances per minute \cite{dingemanse_universal_2015}. Furthermore, there is substantial support for the distinctly human phrase "huh?" being one of the most prevalent linguistic expressions employed, across several language families, to initiate this kind of repair \cite{dingemanse_is_2013} \cite{hayashi_huh_2012}, making it a prime candidate for researchers to investigate when examining interactional language.

\subsection{Interactional Language in HMI}
Previous investigations have primarily focused on interactional language as it pertains to HHI; its application to spoken dialogue systems is relatively new and presents a unique set of challenges and opportunities. Conventional knowledge dictates that SDS strive to emulate human-like interactions \cite{chaves_how_2021}, making it imperative to isolate and analyze the linguistic patterns and conversational norms that implicate this type of communication. This knowledge can inform the design and development of voice assistants, ensuring that they accurately interpret and respond to user queries. 

Exploring interactional language in voice assistants is also vital to improving their usability to the end-user — by identifying areas wherein communication breakdowns arise, researchers can propose enhancements that optimize the interactional design and provide a more intuitive and satisfying user experience. As such, this investigation seeks to determine the significance of interactional language in SDS through an exploration of real-world interactions with popular voice assistant platforms with the aim of identifying their strengths and limitations and proposing insights for future computational linguistics and interactional language research. By bridging these two fields, the present analysis can contribute to ongoing efforts to create SDS capable of navigating the intricacies of human interaction in an informed manner — ultimately paving the way for a more natural, intuitive, and acceptable interactions.

\section{Research Questions / Hypotheses}
\textbf{RQ1: How do Google Assistant and Siri handle dialogue repair when the user is initiating OIR by using “huh?”?}

Given the overall absence of integration of interactional language in SDS, it is likely that these systems will encounter challenges in accurately parsing and handling "huh?" from the user when used as a request for OIR.

\noindent\textbf{RQ2: Is it possible to elicit an OIR strategy like the use of “huh?” when in a dialogue with Google Assistant and Siri?}

It is unlikely that assistants will employ this repair strategy due to the lack of development that the systems have regarding interactional language. Instead, it is assumed that the assistants will rely on more precise and formal language to fulfill the user's request while sacrificing the humanistic quality of interactional language such as "huh?".

\noindent\textbf{RQ3: How do the dialogue repair strategies used by Google Assistant and Siri vary between English and Spanish, if at all?}

Given the limited availability of high-quality Spanish data, it is expected that a discrepancy will arise in the frequency and quality of the strategies employed in English and Spanish across both voice assistants.

\noindent\textbf{RQ4: How acceptable do users find the dialogue repair strategies produced by Google Assistant and Siri?}

Considering the potential impact of the system's repair strategy on the outcomes of user-assistant dialogues, users will strongly favor strategies that prioritize trying to fulfill requests at any cost, rather than strategies that attempt to sidestep or outright ignore the misunderstanding.

\section{Methodology}

To evaluate the proficiency of the voice assistants in generating and responding to the need for dialogue repair, two tests (Task A and Task B, respectively) were performed — each including the OIR strategy triggered by the word “huh?”. Later, a third test (Task C) was conducted by surveying two groups, one in English and one in Spanish, to elicit human acceptability judgements for the repair strategies utilized by Google Assistant and Siri. All dialogues were manually categorized according to 10 dialogue repair strategies (seen in table \ref{tab:table1} and adapted from an earlier version of this research, \cite{galbraith_analysis_2021}) collected from the assistants tested. Notably, this study does not aim to test the automatic speech recognition capabilities of SDS, rather it seeks to explore the strategies that are produced as a result of the SDS' dialogue managers decision making process on how to handle induced misunderstandings.

\begin{table}
\resizebox{\columnwidth}{!}{%
\begin{tabular}{|c|p{2.9 in}|}\hline
\hline
\textbf{Strategy} & \multicolumn{1}{c|}{\textbf{Strategy Description}} \\ \hline
1 & Attempts to fulfill requests given the information, excluding the unintelligible part. \\ \hline
2 & Takes a literal interpretation of the unintelligible part of the utterance and attempts to use the appropriate action to fulfill the request. \\ \hline
3 & Fails to understand or mishears — sometimes proposes an error, may be accompanied by the system asking for the user to ask again. \\ \hline
4 & Not capable of fulfilling requests (sometimes cancels the request due to this inability, or remains completely silent). \\ \hline
5 & Asks for the appropriate information that may include or may not include the unintelligible speech. \\ \hline
6 & Searches the internet for information that it attempts to parse from the unintelligible part of the phrase. \\ \hline
7 & Uses information stored about the user to fulfill requests. \\ \hline
8 & Opens (or proposes) an application (that may or may not be available on the user’s device) that the user can further use to ask the request again to better fulfill it. \\ \hline
9 & Gives the user a list of instructions (or the website to find the information) for the operation of the assistant. \\ \hline
10 & Attempts to phonetically transcribe the unintelligible part of the utterance. \\ \hline
\end{tabular}%
}
\caption{Assistant Dialogue Repair Strategies}
\label{tab:table1}
\end{table}

\subsection{Task A - Assistant Repair Production}
The initial test, "Task A", aimed to answer RQ2/RQ3 by testing the assistants' ability to produce repair initiators by having the user purposefully introduce an unintelligible phrase embedded within a request to create a source for misinterpretation in the dialogue. 

\subsection{Task B - Assistant Repair Comprehension}
The subsequent experiment, "Task B", was conducted with the aim of answering RQ1/RQ3 by testing the assistants' skills with handling repair initiators introduced by the user. This was performed by having the user first begin a query, letting the assistant respond, following up this response with the user asking the repair initiator “huh?”, and recording the assistants’ reaction. This procedure was done to a depth of two turns of the dialogue with the assistant.

\subsection{Task C - Repair User Evaluation}
The final task, "Task C," aimed to answer RQ4 through the creation of two surveys: one in American English and one in Spanish from Spain. The surveys were distributed to 50 participants each (n=50), totaling 100 participants – all of whom were native speakers of their respective languages. These surveys utilized a 5-point Likert scale to record acceptability judgments of dialogue samples from Task A and Task B. The dialogue samples, sourced from assistant responses, included 20 dialogues from each task category, resulting in a total of 40 individual dialogues. Each dialogue repair strategy had two dialogues presented, with one dialogue sourced from each assistant when possible.

\section{Results / Discussion}

\begin{figure}[H]
    \centering
    \captionsetup{justification=centering}
    \includegraphics[width=1\linewidth]{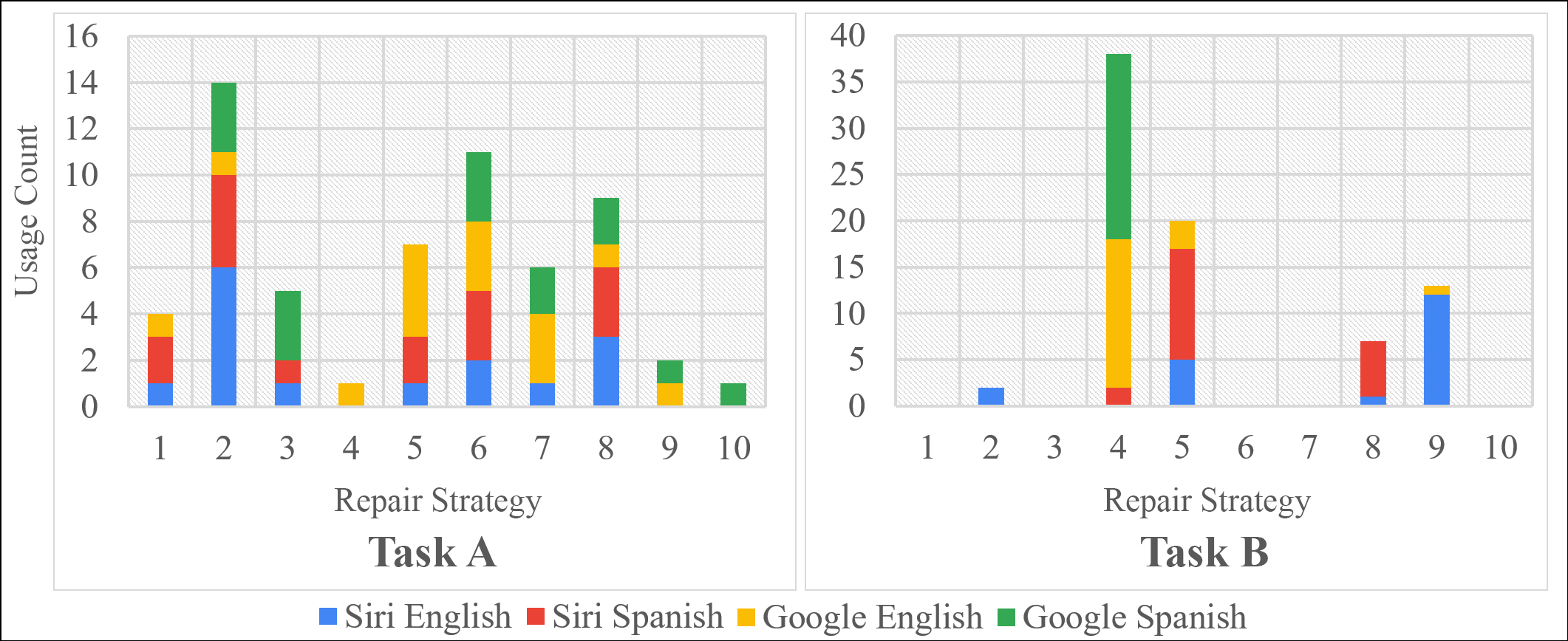}
    \caption{Machine Repair Strategies from Task A and Task B}
    \label{fig:figure1}
\end{figure}

\begin{figure}[H]
    \centering
    \captionsetup{justification=centering}
    \includegraphics[width=1\linewidth]{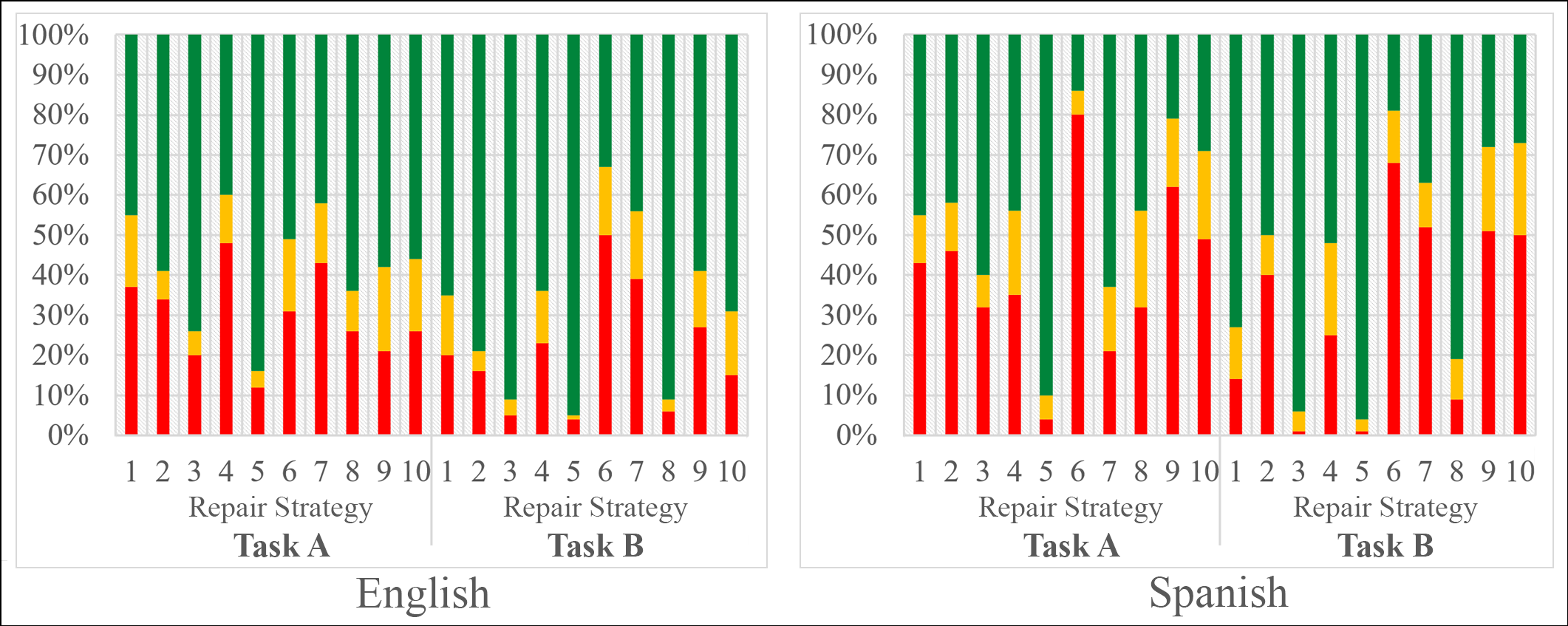}
    \caption{User Acceptability Judgements Across Tasks and Languages}
    \label{fig:figure2}
\end{figure}

While examining whether Google Assistant and Siri can generate an other-initiated repair like the commonly used "huh?" in HHI in Task A (\textbf{RQ2}), findings revealed a discrepancy between these virtual assistants and their human counterparts. Both assistants, in English and Spanish, failed to produce the specific utterance "huh?" when faced with unintelligible input, indicating an inability to replicate this human-like repair strategy (a summary of all repair strategies that the assistants used can be seen in figure \ref{fig:figure1}). Instead, the machines predominantly employed a repair strategy that adopted a literal interpretation of the unintelligible part of the utterance and aimed to fulfill the request accordingly (strategy 2). However, overuse of this strategy contradicted the preferences expressed by human participants, who clearly indicated a preference for a repair attempt that seeks appropriate information, with or without the inclusion of the unintelligible speech (strategy 5) (\textbf{RQ4}) (a summary of all acceptability judgements can be seen in figure \ref{fig:figure2}). 

\begin{figure}[H]
    \centering
    \includegraphics[width=1\linewidth]{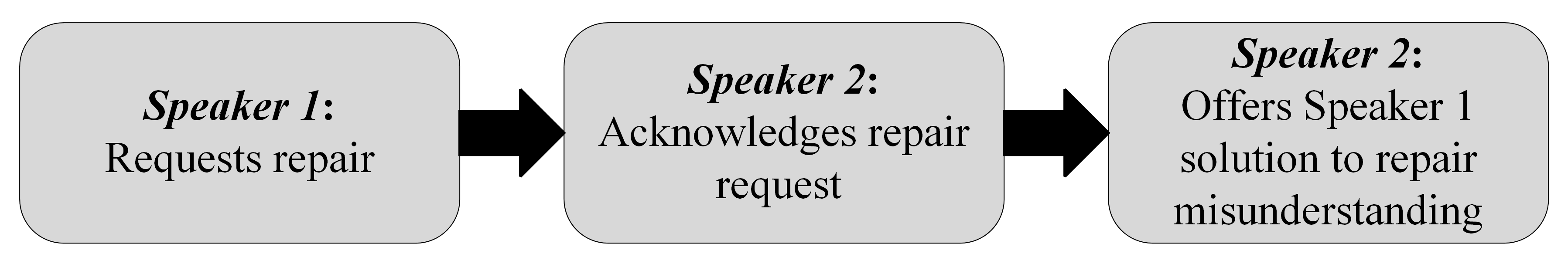}
    \caption{Interactional Paradigm}
    \label{fig:figure3}
\end{figure}

Participants differed slightly in their judgements of other strategies, particularly the widely unacceptable strategy 6 which searches the internet for information to fulfill a request. English speakers tended to perceive this strategy as slightly more acceptable compared to Spanish participants (who universally judged it as completely unacceptable), although it remained the second most unacceptable strategy in English across both tasks combined. Overall, the universal distaste for this repair approach (which defies the hypothesis posited for \textbf{RQ4} which anticipated the least acceptable strategy to be one that outright admitted failure like strategy 4) may be because it does not directly address the repair request in the dialogue like strategy 5, rather it ambiguously addresses the \textit{acknowledgement for the repair request} — the second step within the turn-taking steps of the interactional paradigm seen in figure \ref{fig:figure3}.

Analyzing the response of the virtual assistants to the user's introduction of the interactional word "huh?" in Task B (the right-hand graph in figure \ref{fig:figure1}), both differences and similarities were observed in comparison to how humans employ this word (\textbf{RQ1}). Variations were found in how the assistants handled these interactions across different languages (\textbf{RQ4}), as well as in how English and Spanish speakers evaluated these interactions (\textbf{RQ3}). When tasked with responding to the repair initiator "huh?", the machines employed significantly fewer strategies compared to Task A, utilizing only 5 strategies out of 10. The dominant strategy, primarily utilized by Google Assistant, involved the inability to fulfill the user's request (strategy 4), revealing a complete incapacity to effectively address repairs triggered by "huh?". The second most utilized strategy, strategy 5, was more widely employed by both assistants across the two languages tested. Consistent with the findings of Task A, strategy 5 was deemed the most acceptable by participants (contrary to the assistants' most used strategy in this task), further emphasizing the importance of employing strategies in HMI that uphold a dialogue structure mirroring the turn-taking principles observed in interactional language dialogue repair in HHI (as depicted in figure \ref{fig:figure3}). These qualities are absent in strategy 6, which was the least acceptable not only in this task, but among nearly all tasks and languages overall.

\section{Conclusions}
This study highlights notable disparities and similarities in the utilization of interactional language between HHI and HMI. The findings suggest inconsistencies in the ability of virtual assistants like Google Assistant and Siri to replicate human-like repair strategies, particularly in relation to the use of and reaction to the interactional word "huh?". These results call for further research to explore ways to enhance the ability of virtual assistants to employ repair strategies that align more closely with human interactional language, and to react accordingly when this language is utilized by the user. Infelicities between the assistants' repair strategies and user judgments strengthen this call and also highlight the need for cross-cultural user feedback in future investigations of interactional language in HMI.

\printbibliography

\appendix

\end{document}